# Inference of Upcoming Human Grasp Using EMG During Reach-to-Grasp Movement

Mo Han\*, Mehrshad Zandigohar, Sezen Yağmur Günay, Gunar Schirner and Deniz Erdoğmuş

*Department of Electrical and Computer Engineering, Northeastern University, Boston, MA, United States*

Electromyography (EMG) data has been extensively adopted as an intuitive interface for instructing human-robot collaboration. A major challenge to the real-time detection of human grasp intent is the identification of dynamic EMG from hand movements. Previous studies predominantly implemented the steady-state EMG classification with a small number of grasp patterns in dynamic situations, which are insufficient to generate differentiated control regarding the variation of muscular activity in practice. In order to better detect dynamic movements, more EMG variability could be integrated into the model. However, only limited research was conducted on such detection of dynamic grasp motions, and most existing assessments on non-static EMG classification either require supervised ground-truth timestamps of the movement status or only contain limited kinematic variations. In this study, we propose a framework for classifying dynamic EMG signals into gestures and examine the impact of different movement phases, using an unsupervised method to segment and label the action transitions. We collected and utilized data from large gesture vocabularies with multiple dynamic actions to encode the transitions from one grasp intent to another based on natural sequences of human grasp movements. The classifier for identifying the gesture label was constructed afterward based on the dynamic EMG signal, with no supervised annotation of kinematic movements required. Finally, we evaluated the performances of several training strategies using EMG data from different movement phases and explored the information revealed from each phase. All experiments were evaluated in a real-time style with the performance transitions presented over time.

## 1. INTRODUCTION

With the rapid development of human–robot interaction (HRI) technology, collaborative robotics have been widely utilized in the assistive environment and smart prosthetic hands. The activity detection of the human hand and arm (Sheikholeslami et al., 2017; Zandigohar et al., 2019; Han et al., 2021) is an intuitive interface for instructing the cognitive collaboration between humans and robots without requiring users to have professional control skills.

To extract the motion instructions, electromyography (EMG) signal collected from arm and hand has been extensively adopted since it can accurately detect the motion intention and does not require invasive data collection (Ju and Liu, 2013; Han et al., 2019, 2020; Zangigohar et al., 2021).

Online human–robot interaction through hand and arm motion is hard to achieve due to the high degrees of freedom (DOFs) of human body structure. More specifically, the human hand alone consists of 21 DOFs controlled by 29 muscles (Jones and Lederman, 2006). Most previous studies focused on investigating the discrete classifications of hand and arm movements, by exploiting the mapping between EMG signals and hand postures (Sebelius et al., 2005; Dalley et al., 2011; Ju and Liu, 2013; Ouyang et al., 2013; Han et al., 2019, 2020). However, this is insufficient to generate differentiated control regarding the variety of practical dynamic EMG signals. In real-world applications, the muscular activity varies between a static and a dynamic arm position, and the hand configuration also changes simultaneously with the arm motion (Jiang et al., 2013). Moreover, those studies have only considered a small number of grasp patterns, which cannot ensure the model robustness as required by the grasping of a larger variety of objects. To improve the control effectiveness and user comfort, the human intention should be detected in a more dynamic, natural, and smooth manner.

In order to increase the system's applicability to a wider range of movements, one could integrate more EMG variability into the model training and validation using transient EMG signals from different dynamic phases (Yang et al., 2012; Castellini et al., 2014). Furthermore, the hand motions are commonly carried out in concert with the dynamic movements of the arm. For example, when the hand is approaching a target object to be grasped, the configuration of the fingers and wrist also changes simultaneously during the reach-to-grasp motion according to the shape and distance of the object (Jeannerod, 1984). Therefore, the identification of varying muscular contractions and dynamic arm postures could provide more response time for pre-shaping the robot, which could improve the system usability and result in more natural grasp transitions.

However, only limited research was focused on detecting dynamic grasp motions (Lorrain et al., 2011; Siu et al., 2016; Batzianoulis et al., 2017; Sburlea and Müller-Putz, 2018). Among those studies, in (Sburlea and Müller-Putz, 2018), a multimodal dataset was acquired, which consists of electroencephalographic (EEG), kinematic, and EMG recordings performing multiple grasp types in three dynamic stages of the grasping movement. However, the EMG data were recorded by the Myo armband, which only contains eight EMG channels surrounding the lower arm and, thus, cannot provide adequate higher-arm movement indication. This results in a less adaptive amputation level and a limited identification of the reaching movement of the arm. In addition, the classification of EMG data acquired from dynamic movements was not fully discussed in this study. Researchers in (Batzianoulis et al., 2017) proposed an EMG-based learning approach to decode dynamic reach-to-grasp movements by measuring the ground-truth finger configurations with a wired glove. The authors of (Siu et al., 2016) also explored a classifier to identify transient anticipatory EMG signals incorporating dynamic grasp actions, assisted by a kinematic device detecting the ground-truth timestamps of action transitions. In (Lorrain et al., 2011), the grasp classification was evaluated on data involving both static and dynamic contractions, but the data collection experiments were conducted in a discrete manner, where the subjects were required to maintain the grasping and resting positions alternately for specified time lengths, so the practical movement continuity was still missing. In other words, most existing assessments on non-static EMG classification either require the supervised ground-truth timestamps of the movement status or only contain limited information and dynamic variations of the EMG signal.

Therefore, we propose a framework for classifying dynamic EMG signals into gestures and examine the impact of different phases of reach-to-grasp movements on final performance. We exploit the continuity of hand formation change to increase data variability and decode the subject's grasping intention in a real-time manner. We utilized EMG data from large vocabularies of gestures with multiple dynamic motion phases, which enabled us to encode the transitions from one intent to another based on natural sequences of human reach-to-grasp movements. During the data collection, continuous variations on multi-scale muscular contractions were introduced by the designed experiment protocol simulating the actual situation. We segmented the continuous EMG data unsupervised into different dynamic motion sequences, and further labeled those EMG sequences according to the specific motion. The classifier for identifying gesture labels was constructed afterward based on the dynamic EMG signals, with no supervised annotation of kinematic movements required. Finally, we examined the performances of several training strategies using EMG data from different dynamic phases and explored the information revealed from each phase. All experiments were evaluated in a real-time style with the performance variation presented over time. The proposed method was shown to be efficient due to the greater amount of information introduced by dynamic motions.

## 2. MATERIALS AND METHODS

### 2.1. Experimental Protocol and Data Processing

The utilized data were collected from 5 healthy subjects (4 men, 1 women; mean age: 26.7 ± 3.5 years). All subjects were right-handed and only the dominant hand was used for data collection. None of the subjects had any known motor or psychological disorders. The experimental procedure and tasks were explained to all subjects, and we obtained their consent to participate before the experiments.

#### 2.1.1. EMG Sensor Configurations

We collected surface EMG (Motion Lab Systems, Baton Rouge, LA, USA) in bipolar derivations, with a sampling rate of $f = 1562.5$ Hz. The visualization of the $C = 12$ targeted muscles of our experiment is shown in **Figure 1**, including muscles ranging from hand to upper arm in order to capture more dynamic movement information. The 12 muscles are: First Dorsal Interosseous (FDI), Abductor Pollicis Brevis (APB), Flexor Digiti Minimi (FDM), Extensor Indicis (EI), Extensor Digitorum Communis (EDC), Flexor Digitorum Superficialis (FDS), Brachioradialis (BRD), Extensor Carpi Radialis (ECR), Extensor Carpi Ulnaris (ECU), Flexor Carpi Ulnaris (FCU), Biceps Brachii Long Head (BIC), and Triceps Brachii Lateral/Short Head (TRI).

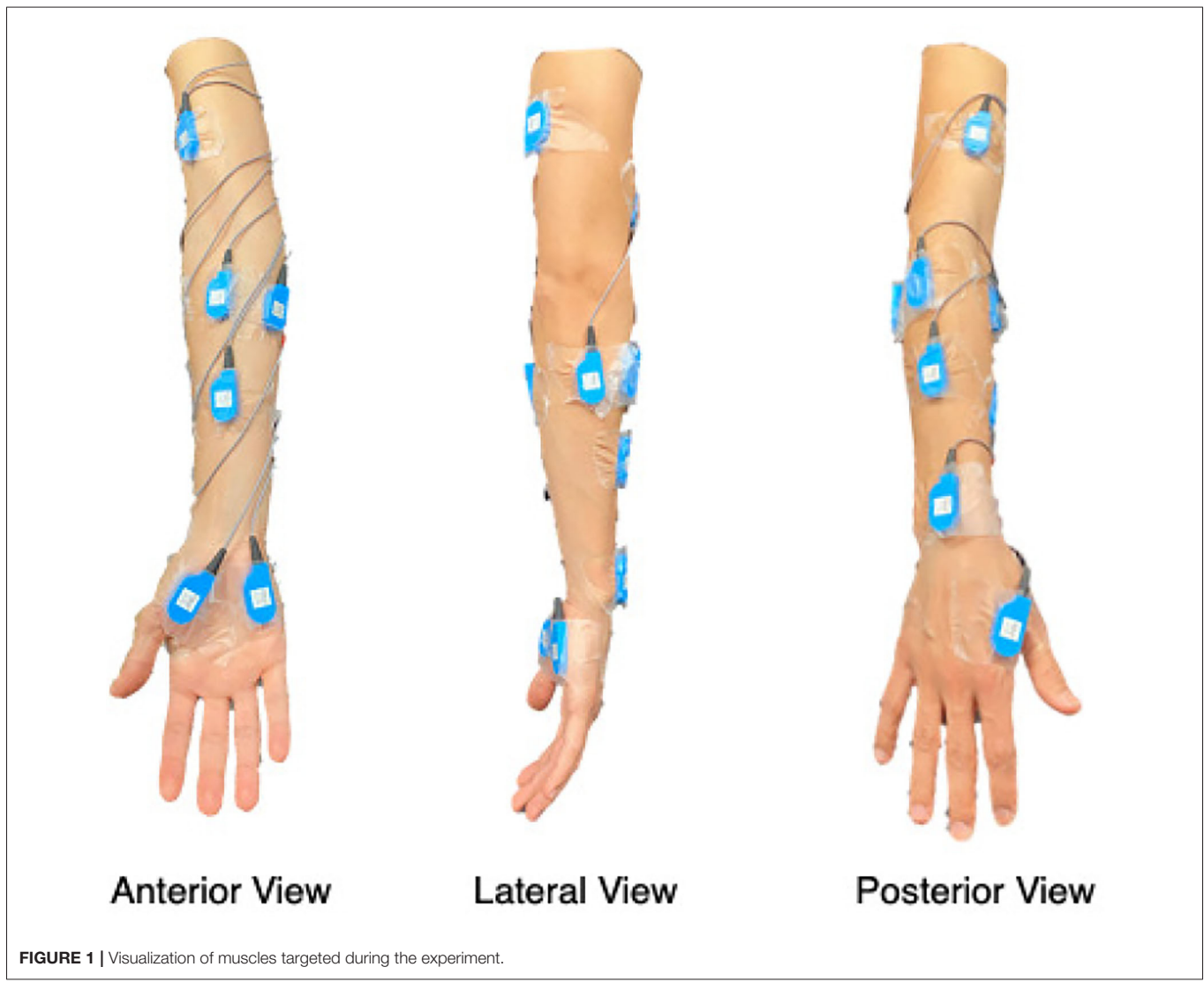

**FIGURE 1** | Visualization of muscles targeted during the experiment.

### 2.1.2. Experimental Protocol

The experimental protocol focused on 14 gestures and 4 dynamic motion phases involving commonly used gestures and wrist motions (Feix et al., 2016). As shown in **Figure 2**, the 14 classes were: large diameter, small diameter, medium wrap, parallel extension, distal, tip pinch, precision disk, precision sphere, fixed hook, palmar, lateral, lateral tripod, writing tripod, and open palm/rest. We defined the 14 grasp labels as $l \in \{0, 1, ..., 13\}$, where $l = 0$ indicated the open-palm/rest gesture without any movement, and $l \in \{1, ..., 13\}$ were accordingly identified as the other 13 gestures listed in **Figure 2**.

Each subject participated in two collection sessions in total, involving the task of lifting and moving different objects from one position to another, where in the first session the object was moved in a clockwise trajectory while the second session was in a counterclockwise trajectory. The subjects were asked to rest for 15 min between the two sessions.

During the sessions, each gesture of $l \in \{1, ..., 13\}$ (not including the open-palm/rest gesture) was performed four times using four different objects, totaling 52 objects in each session. The subject performed 6 trials for each of the 52 objects per session, where each trial was executed along its corresponding predefined path, as shown in **Figure 3**. During the first trial t1, the object was moved from the initial position P0 to the position P1, followed by another five trials to move the object clockwise until it was returned to the initial position P0. In each trial to move the target object from one position to another (e.g., from P0 to P1), the hand first reached the current position of the object, then grasped the object and moved it to the subsequent position, and next to the hand moved back to the rest position. The counterclockwise session was performed in a similar manner as **Figure 3** but in a different direction with respect to the initial position P0. Therefore, the grasping movement could be conducted from different angles, directions, and distances toward the target object, which increases the variety of the collected dynamic EMG data.

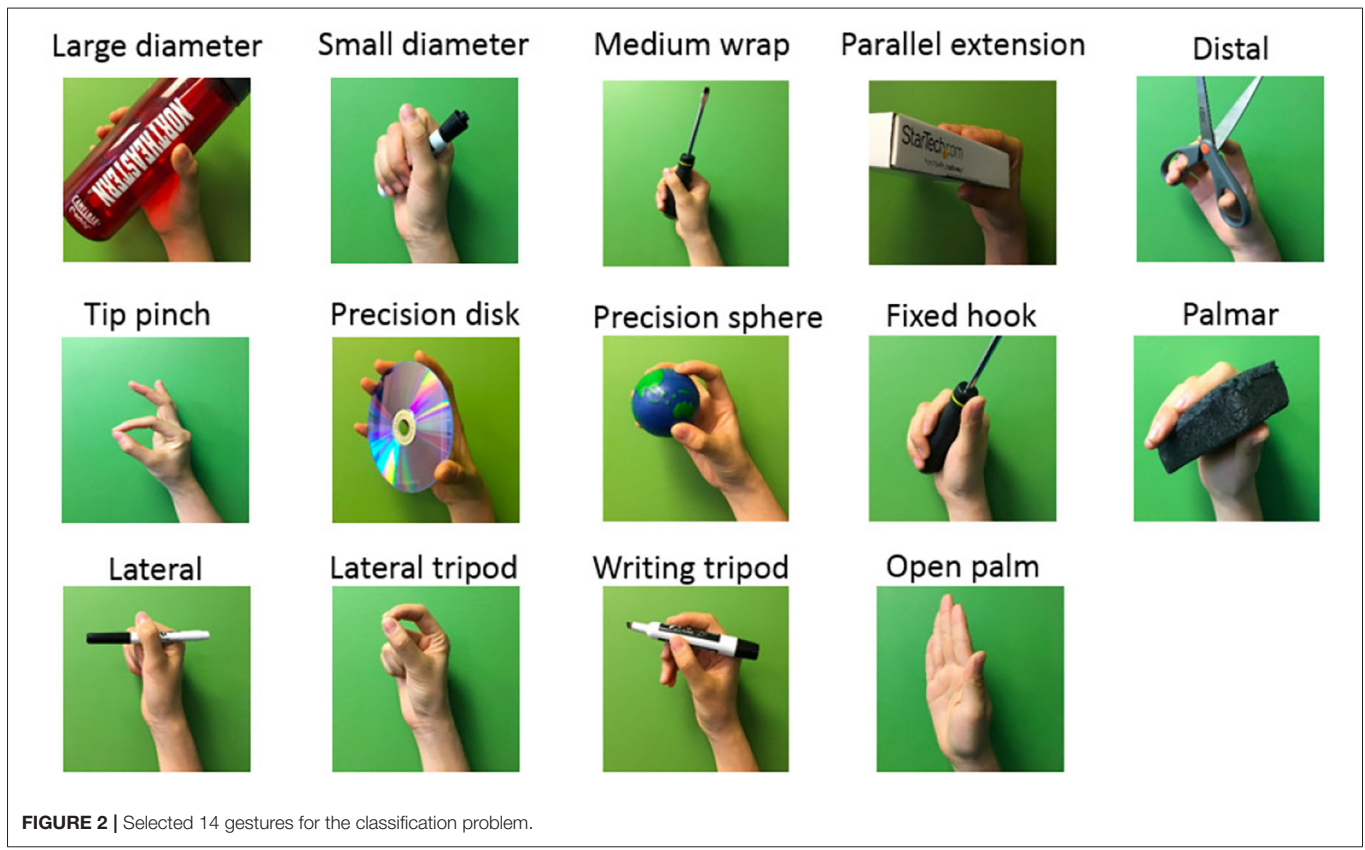

FIGURE 2 | Selected 14 gestures for the classification problem.

At the beginning of the experiment, the subject was seated facing a table and electrodes were connected to the right arm while the arm was at the rest position with an open palm, as illustrated in **Figure 3**. Object center configuration was defined with 6 marks on the table. A screen was placed on the right side of the subject for showing example pictures of the gestures to be executed. First, the subject was given 5 s to read the gesture shown on the screen, followed by an audio cue illustrating the beginning of the first trial. Each trial lasted for 4 s, and the object was grasped and moved along its predefined path using the designated gesture for 6 trials without interruption, with audio cues given between different trials. Within each 4-s trial, the subject was required to complete 4 actions that could be naturally performed by a human during reach-to-grasp movements, including: (1). reaching (reaching the object), (2). grasping (grasping to move the object), (3). returning (returning to the rest position), and (4). resting (resting at the rest position with an open palm). The complete timeline for grasping each object is presented in **Figure 4**. Note that for each trial, in order to preserve sufficient information on the dynamic motion, the four grasp phases (reaching, grasping, returning, and resting) were performed freely and naturally by the subject without limitation on the speed of each phase as long as all the 4 phases were completed within 4 s.

### 2.1.3. Data Pre-processing

Since the experimental procedure requires subjects to execute real movements, the data is prone to noises and motion artifacts. Thus, we applied a fourth-order band-pass (40 Hz to 500 Hz). Butterworth filter to remove the unwanted data contamination and clear any other frequency noise outside of the normal EMG range. No default filtering was applied to the data from the acquisition device.

Due to the nature of the human muscle system, upper arm muscles generate stronger signals than hand muscles and thus a fair source contribution would be only possible by normalizing each muscle with respect to its maximum power. Therefore, the maximum voluntary contraction (MVC) test was conducted for each muscle where the subjects were asked to perform isometric constructions of muscles lasting for 3 s. The envelope of both the experiment and the MVC data were generated, and each channel of the experimental data was then normalized with respect to the maximum value of MVC envelopes.

In order to implement the EMG identification in real-time, we further divided the processed experimental data of EMG

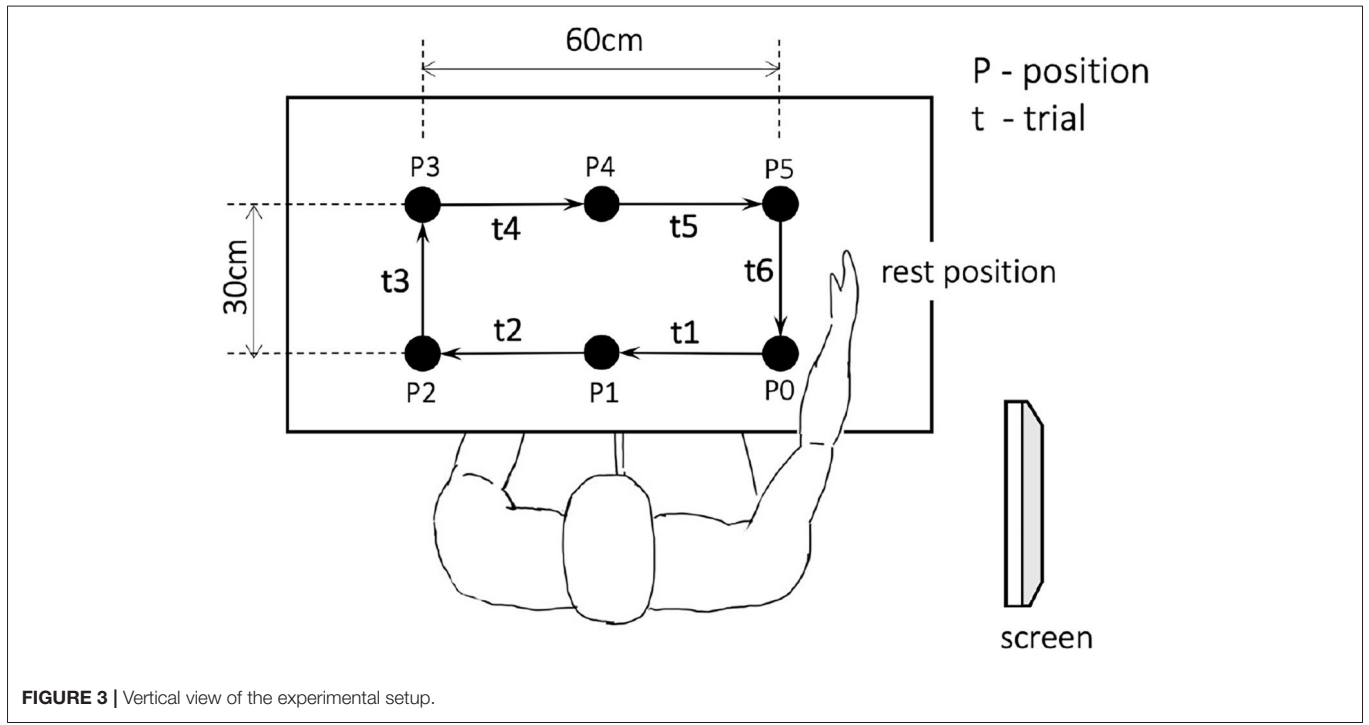

FIGURE 3 | Vertical view of the experimental setup.

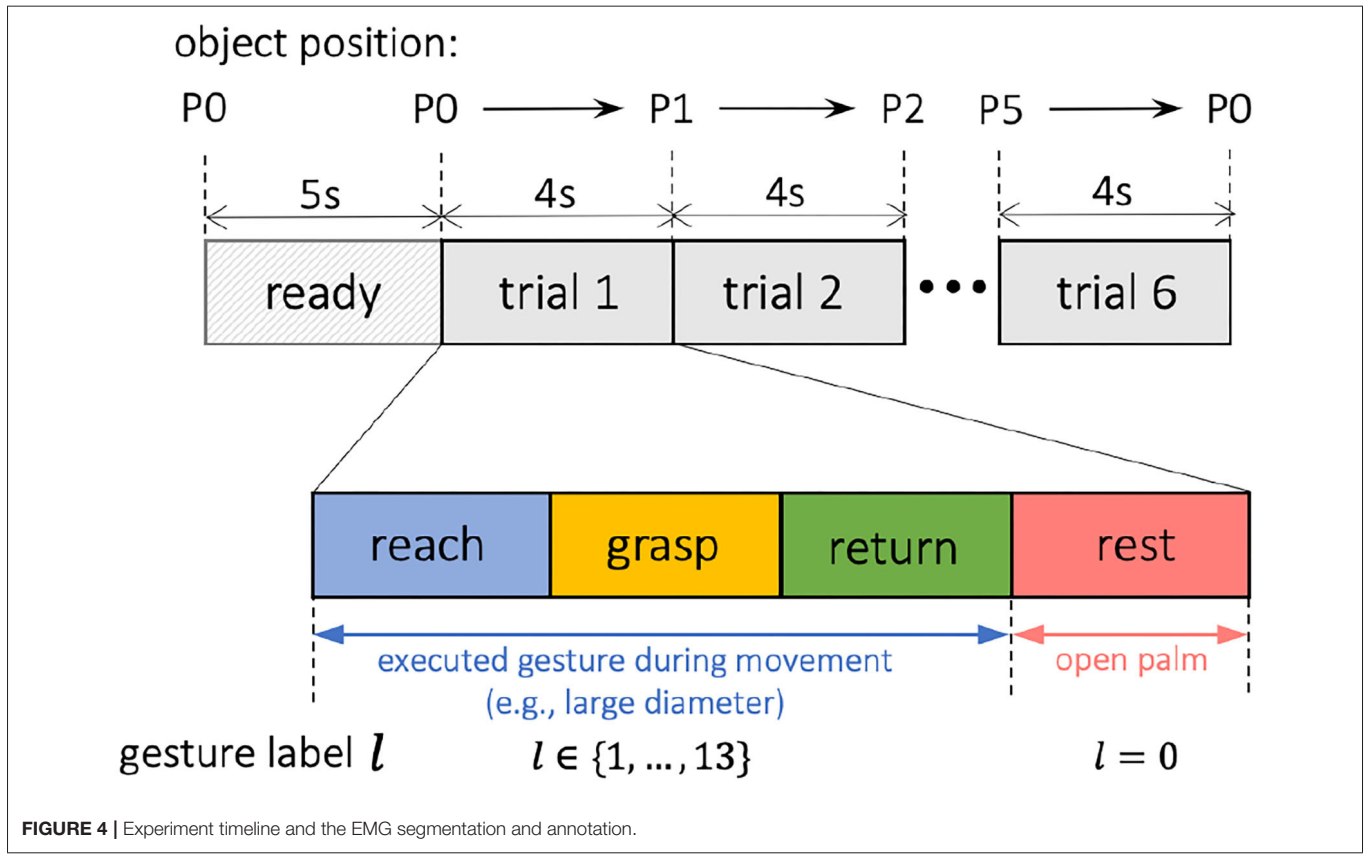

FIGURE 4 | Experiment timeline and the EMG segmentation and annotation.

envelopes into sliding time windows of $T = 320$ ms, with a step size of 40 ms between two consecutive windows. The following EMG feature extraction and classification were all conducted based on each time window.

## 2.2. Methodologies
### 2.2.1. Feature Extraction

The selection of EMG features involves considering both processing time and data representability, at the same time avoiding redundancy to maximize the classification performance (Phinyomark et al., 2012). The time-domain features require lower computational complexity compared to frequency-domain features and thus can be implemented in real-time with higher speed (Phinyomark et al., 2012). In this study, three time domain features were extracted, including root mean square (RMS), mean absolute value (MAV), and variance of EMG (VAR). As illustrated in Hogan and Mann (1980), the maximum likelihood estimation of EMG amplitude can be evaluated by the RMS feature, since under constant-force, constant-angle, and non-fatiguing construction, EMG signals can be modeled as Gaussian distributions. Additionally, MAV is a common feature for indicating EMG amplitude due to its low computational requirements and potential for higher class distinction (Phinyomark et al., 2010, 2011). Finally, previous experimental studies also revealed that the VAR feature could improve the EMG classification performance, which is another frequently used feature in EMG classification studies (Phinyomark et al., 2010, 2012).

The feature extraction was then applied to the pre-processed EMG time windows, and RMS, MAV, and VAR features were calculated over the window input of $X \in \mathbb{R}^{C \times T}$, where $C = 12$ is the channel number of EMG from all muscles and $T = 320$ ms is the window length with a sampling rate of $f = 1562.5$ Hz. Then the output feature vector of $Z \in \mathbb{R}^{3C \times 1}$ was yielded corresponding to each input window $X \in \mathbb{R}^{C \times T}$.

### 2.2.2. Unsupervised Segmentation of Dynamic Motion

In each trial, the grasp movements were performed naturally by the subject without limitation on the timing of each motion phase. Since the distance between hand and object varied across different trials, the duration of EMG sequences from different motion phases was also not constant. For example, as shown in **Figure 3**, since the object was closer to the hand during t1, the reaching distance of trial t1 was shorter than that of t3. Therefore, to approach the gesture classification in a continuous manner, we first segmented each EMG trial into different movement sequences unsupervised, and then labeled those sequences separately according to the specific motion, as shown in **Figure 4**.

To segment each EMG trial into multiple sequences unsupervised based on the dynamic grasp movements, the method of Greedy Gaussian Segmentation (GGS) by Hallac et al. (2019) was adopted. The GGS algorithm works under the assumption that during a particular static state, the EMG signal can be well explained as a Gaussian random process with zero mean (Clancy and Hogan, 1999). This GGS method aims to break down multivariate time series into several segments, where the observed data in different segments can be well modeled as separate Gaussian distributions which are independent of the other segments. Therefore, the mean and covariance of the Gaussian distribution in each segment are also assumed to be unrelated to the other segments. This time-series segmentation task can then be transformed into a maximum likelihood problem, where the optimal solution is a set of breakpoints maximizing the overall likelihood when sampling from all of the independent Gaussian segments. To decrease the computation time, GGS utilizes a greedy and scalable implementation of dynamic programming where optimal breakpoints are calculated iteratively and re-adjusted in each iteration.

In this research, we segmented all EMG trials using GGS by assigning 3 breakpoints to each trial, which gave us 4 sequences corresponding to reaching, grasping, returning, and resting. By doing so, the order of different movement phases was taken into account by imposing a higher probability of movement transitions that were expected to follow one another, e.g., grasping movement following reaching movement, later followed by returning action. Given the specified number of breakpoints, the GGS algorithm was utilized to unsupervised segment the EMG trial and provide the optimal segment boundaries.

In **Figure 5**, such utilization of the GGS method is depicted, where each segment of the 12-channel EMG series is modeled as an independent multivariate Gaussian distribution with distinct mean and covariance parameters. During the data collection, an eye-tracker was also worn by the subject, where experiment videos synchronized to the EMG data were recorded by the forward-facing world camera of the eye-tracker. In order to validate the effectiveness of the unsupervised segmentation, for each trial, we extracted and reviewed three key video frames corresponding to the three segment boundaries. This makes sure the hand movements shown in the frames are consistent with the motion phase transitions. An example of such experiment key frames is displayed in **Figure 5**, where the hand just reaches the object without lifting it at the first break point, the hand releases the object and starts returning at the second break point, and finally, the hand returns to the rest position at the last break point.

### 2.2.3. Dynamic Hand Gesture Annotation

Throughout the reach-to-grasp movement, the limb configurations of fingers and wrist together change continuously with respect to the shape and distance of the targeted object (Jeannerod, 1984). A closer look would reveal the fact that humans tend to pre-shape our hands prior to touching the targeted object. In order to recognize these hand gestures based on the upcoming dynamic data, we annotated the collected EMG with a set of gesture labels $l \in \{0, 1, ..., 13\}$, where $l = 0$ was defined as the open-palm/rest gesture and $l \in \{1, ..., 13\}$ were accordingly identified as the other 13 gestures listed in **Figure 2**. Based on this definition, following the segmented EMG trials as shown in **Figure 4**, we labeled the three motional EMG sequences of reaching, grasping, and returning to be the gesture $l \in \{1, ..., 13\}$ executed during the movement, and tagged the stationary phase of resting with the open-palm label $l = 0$.

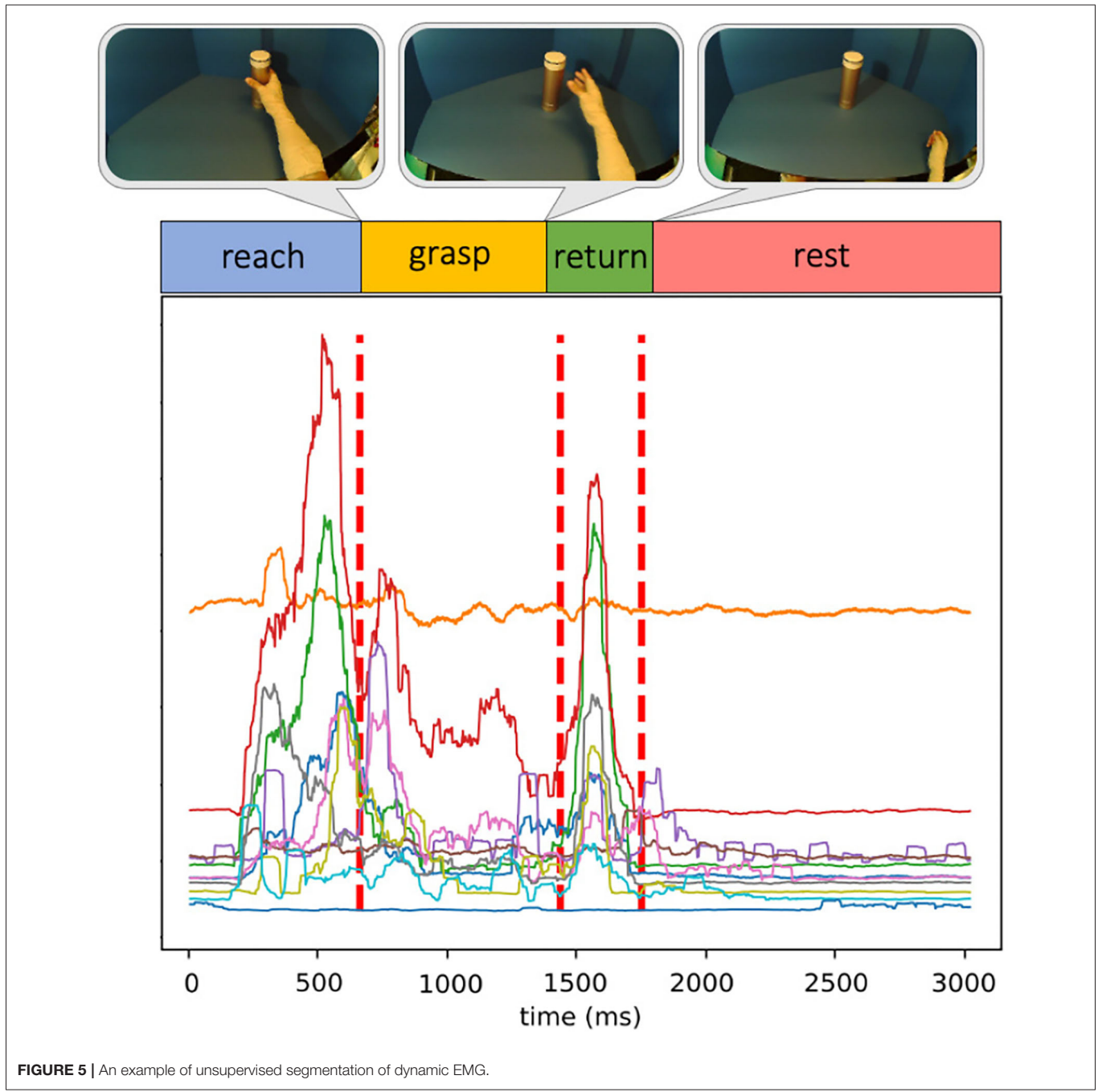

FIGURE 5 | An example of unsupervised segmentation of dynamic EMG.

### 2.2.4. Classification of the Dynamic Hand Gesture

We constructed a classifier to recognize dynamic hand gestures using the collected EMG signals. The EMG signals were first broken down into time windows of $X_i \in \mathbb{R}^{C \times T}$ with size $T = 320$ ms ($f = 1562.5$ Hz sampling rate), where $X_i$ represents the $i$th EMG window and $C = 12$ is the number of channels. Then, based on the gesture label $l \in \{0, 1, ..., 13\}$ of the corresponding EMG window $X_i$, pairs of data and label $\{(X_i, l)\}_{i=1}^n$ were formed, where $n$ is the total number of windows. Later, the selected three time-domain features of RMS, MAV, and VAR were extracted as $Z_i \in \mathbb{R}^{3C \times 1}$ for each EMG window $X_i \in \mathbb{R}^{C \times T}$, leading to feature-label pairs of $\{(Z_i, l)\}_{i=1}^n$.

In this research, we utilized the extra-trees method (Geurts et al., 2006) for identifying hand gestures based on the extracted EMG features. This method incorporates averaging an ensemble of random decision trees trained on different sub-samples of the dataset, which reduces overfitting and improves performance. We observed that the utilization of the extra-trees algorithm with a combination of 50 trees provided desirable performance, hence used in this study.

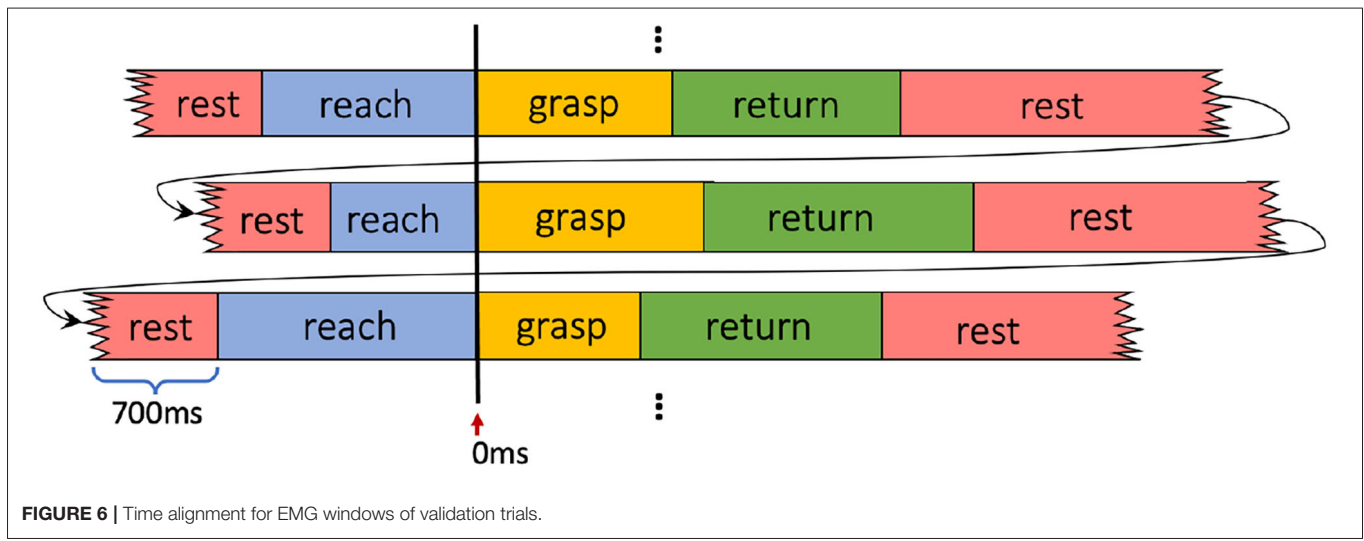

FIGURE 6 | Time alignment for EMG windows of validation trials.

### 2.2.5. Training and Validation

We performed intra-subject training and validation of the 14-class gesture classification, i.e., the training and validation were performed for different subjects separately. For every subject, we implemented a 3-fold validation protocol. For each validation fold, we randomly split the 6 trials collected from each object into 4 training trials and 2 validation trials, totaling 224 trials (66.7%) for training and 112 trials (33.3%) for validation in total. Thereafter, the classifier was trained on the aforementioned training set and evaluated on the left-out validation set which was unseen to the classifier. Since our main objective was to detect the upcoming grasping intention at an earlier stage and pre-shape the robot before the final grasp was accomplished, the data from the returning phase were excluded from training, during which the grasp was already completed and the object was already released from the hand. However, to provide a full scope of the effectiveness of the classification approach, the classifier was then evaluated on the entire EMG trial of four movement phases irrespectively.

As the subjects were not required to perform the four movement phases of reach-grasp-return-rest at a fixed timing or speed during the data collection, the duration of each phase for different trials varies based on the subject's natural pace. Therefore, during the evaluation, we aligned all validation EMG trials based on the breakpoint occurring at the beginning of the grasping phase, as shown at 0 ms in **Figure 6**. In this way, the evaluation could focus more on the assessment of the transition from reaching to grasping, which is the key transition in grasp intent inference. Thus, the performances across all the validation trials can be integrated more clearly by averaging them based on their aligned timelines. In the last step, the resting-phase data of 700 ms from the end of the previous trial was appended at the beginning of its following trial in order to show the dynamic transition between resting and reaching phases.

### 2.2.6. Classification Approach

As mentioned earlier, we trained the classifier with reaching, grasping, and resting phases only, whereas validated the model with all motion phases (reaching, grasping, returning, and resting). In order to specifically inspect the performance of the dynamic-EMG classifier and explore the informative levels of EMG data from different motion phases, the constructed gesture classifier was trained with three different strategies independently:

1. Trained with EMG of reaching and resting phases;
2. Trained with EMG of grasping and resting phases;
3. Trained with EMG of reaching, grasping, and resting phases.

## 3. RESULTS

The performances of classifiers trained by the three different strategies introduced in Section 2.2.6 are shown in **Figures 7**–**9**, respectively. We present the validation results as functions of time, in order to inspect the performance variation during different dynamic phases during a trial. In those figures, we show the predicted probabilities and the classification accuracy on the validation set, where each time point represents an EMG time window. The performances of each time point in a trial were averaged over all 3-fold validation trials of all subjects, given their aligned timeline. We define the beginning of the grasping phase as 0 ms, and the averaged breakpoints between different motion phases are illustrated by vertical dashed lines.

In **Figures 7**–**9**, the *grasp gesture* is identified as the ground-truth gesture $l \in \{1, ..., 13\}$ executed during the non-resting phases, the *rest gesture* represents the open-palm/rest gesture $l = 0$ during the resting phase, and the *top competitor* is defined as the gesture with the highest predicted probability among all other 12 gestures which were not performed during the entire trial. For example, a large diameter was executed in order to grasp a bottle during a trial, so in this case, the *grasp gesture* is the large diameter with $l = 1$ and the *rest gesture* denotes open-palm/rest gesture $l = 0$. Since the classification was applied to each EMG time window independently, every time point $t$ in **Figures 7**–**9** corresponds to a 14-dimensional predicted probability output $(p_{t,l=0}, p_{t,l=1}, ..., p_{t,l=14})$, where the *rest gesture* probability is $p_{l=0}$, the *grasp gesture* probability is $p_{l=1}$ (given the large diameter as *grasp gesture*), and the *top competitor* probability is $\max_{j \neq 0,1} \{p_{t,l=j}\}$. Note that for different time points $t_m$ and $t_n$, the *top competitor* gesture $\max_{j \neq 0,1} \{p_{t_m,l=j}\}$ and $\max_{j \neq 0,1} \{p_{t_n,l=j}\}$ can be distinct so the *top competitor* does not represent a particular grasp type and varies over time windows. The probability distance between the *grasp gesture* and the *top competitor* indicates the classifier's capability of distinguishing ground-truth gestures from other distracting grasps. The predicted probabilities and classification accuracies in **Figures 7**–**9** were averaged over all subjects and validation trials, i.e., the classification performance was evaluated and averaged on all possible *grasp gestures* $l \in \{1, ..., 13\}$, not referring to the performance on a specific gesture.

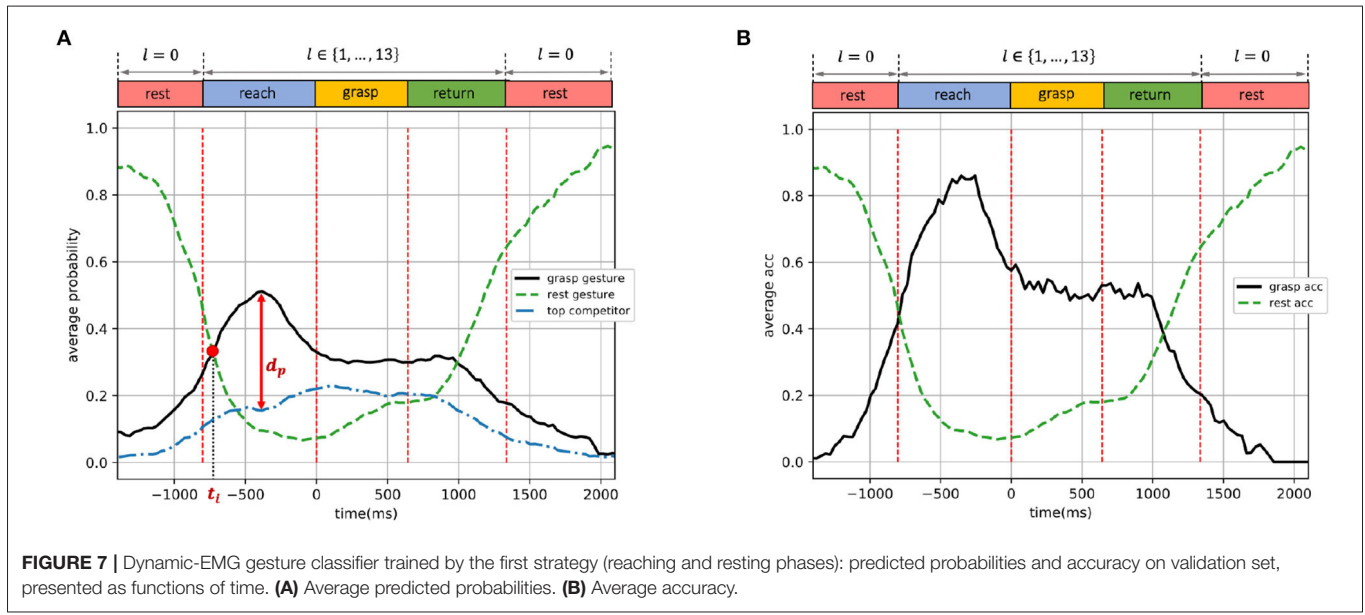

FIGURE 7 | Dynamic-EMG gesture classifier trained by the first strategy (reaching and resting phases): predicted probabilities and accuracy on validation set, presented as functions of time. (A) Average predicted probabilities. (B) Average accuracy.

In **Figures 7, 8, 9A**, the time point at which the probabilities of *grasp gesture* and *rest gesture* intersect with each other is defined as $t_i$ (the intersection is marked by a solid red dot), and the distance from the probability peak of *grasp gesture* to the simultaneous probability of *top competitor* is defined as $d_p$. In **Figures 7, 8, 9B**, the accuracies of successfully detecting the *grasp gesture* and *rest gesture* are independently given.

In this section, the performances of the three training strategies are further compared and analyzed, and the information revealed by dynamic EMG of different motion phases is also discussed.

## 3.1. Training With Reaching and Resting Phases

The performance of the classifier trained by the first strategy (trained with reaching and resting phases) is shown in **Figure 7**. As illustrated in **Figure 7A**, since the classifier was trained only by the reaching phase, not including the grasping phase, the peak of the predicted *grasp gesture* probability was achieved during the reaching phase at around 0.51. During the grasping phase and the first half of the returning phase, the *grasp gesture* probability was stable at around 0.3, and then gradually decreased as the movement converted to the resting phase. Notably, the estimated probability of *grasp gesture* was constantly higher than the *top competitor* throughout the entire trial. The time point when the probability curves of *grasp gesture* and *rest gesture* intersect was $t_i = -729$ ms and the distance from the probability peak of *grasp gesture* to the *top competitor* probability was $d_p = 0.36$. Simultaneously, the predicted probability of the *rest gesture* reduced dramatically to below 0.2 as the grasp movement happened until the hand returned to the rest position when the open-palm probability gradually went up again.

In **Figure 7B**, the average accuracy for detecting *grasp gesture* during the reaching phase presented an outstanding performance with a peak over 0.8 and remained higher than 0.6 within most of the reaching phase. This accuracy decreased in grasping and returning phases, fluctuating around 0.5. The performance of detecting *rest gesture* was highly accurate, with the accuracy value over 0.8 throughout most of the two resting phases.

We are especially interested in the intersection point $t_i$ between the *grasp gesture* and *rest gesture* probabilities. Since after this point, the *grasp gesture* starts to outperform the *rest gesture*, the system could start to prepare for pre-shaping the robot from this point on according to the type of detected *grasp gesture*. Ideally, this intersection is expected to appear right at the junction where the resting phase ends and the reaching phase starts in order to indicate the beginning of the hand motion. However, in practice, the hand movement could only be predicted based on the past motion, so the intersection is expected to appear after the start of the reaching phase but the closer to it the better. In **Figure 7A**, the intersection of the two curves appeared $t_i = -729$ ms earlier than the start of the grasping phase, which was after but very close to the beginning of the reaching phase and allowed enough time to pre-shape the robotic hand before the actual grasp. Therefore, the dynamic EMG of the reaching phase is proved to be informative for forecasting the gesture in advance during the reach-to-grasp movement. Furthermore, even though the model was trained by reaching and resting phases only, the probability of *grasp gesture* was still steadily higher than *top competitor* during all the four dynamic phases, representing the significant hand-shaping information revealed by the movement during the reaching phase.

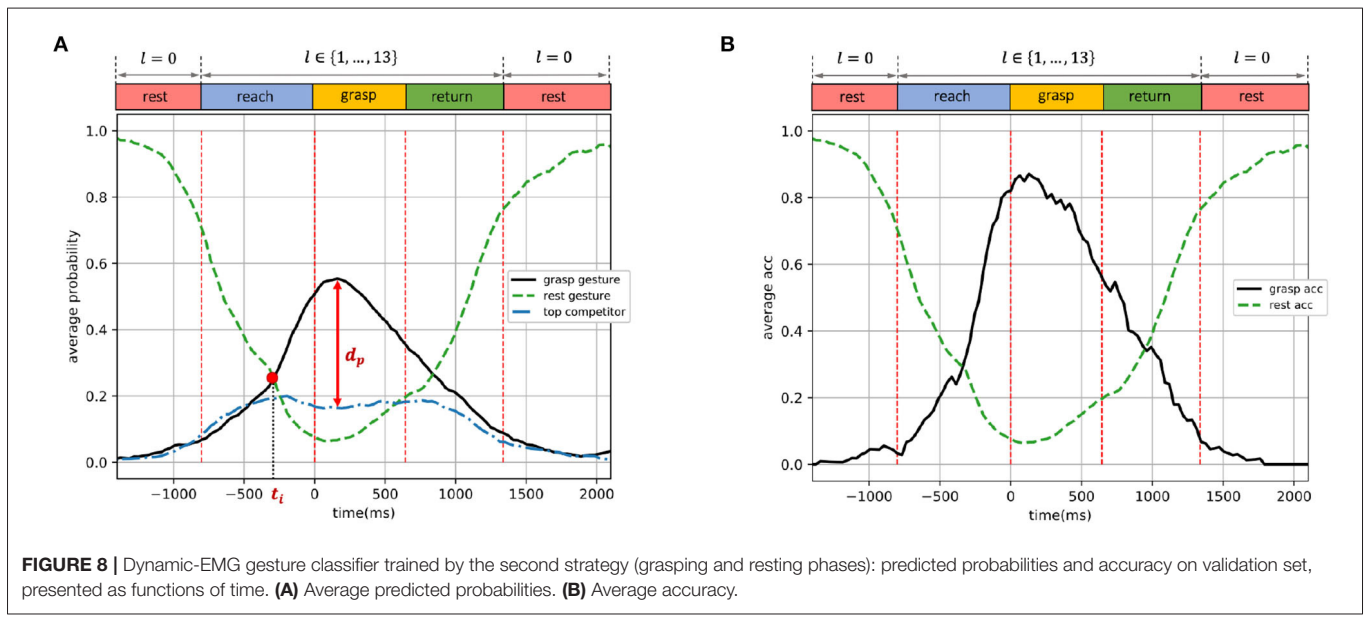

**FIGURE 8** | Dynamic-EMG gesture classifier trained by the second strategy (grasping and resting phases): predicted probabilities and accuracy on validation set, presented as functions of time. **(A)** Average predicted probabilities. **(B)** Average accuracy.

## 3.2. Training With Grasping and Resting Phases

The performance of the classifier trained by the second strategy (trained with grasping and resting phases) is shown in **Figure 8**. As shown in **Figure 8A**, the predicted *grasp gesture* probability first gradually increased, until the grasping phase where the probability peak of *grasp gesture* was reached at around 0.56. When the hand movement entered returning and resting phases, the *grasp gesture* probability declined to lower than 0.2. The probability curve of *rest gesture* shows an opposite trend—the estimated *rest gesture* probability started from a peak at the first resting phase, then decreased to below 0.2 as the grasp movement happened, and then finally reached another peak when the hand went back to rest again. The intersection point between *grasp gesture* and *rest gesture* probabilities appeared at $t_i = -334$ ms, and the probability peak of *grasp gesture* was $d_p = 0.41$ higher than the *top competitor* probability at the same time point.

In **Figure 8B**, the average accuracy of detecting *grasp gesture* during the grasping phase performed outstandingly better than other phases, showing a similar pattern with its probability curve in **Figure 8A**. This *grasp gesture* accuracy was also higher than 0.6 during the end of the reaching phase and the entire grasping phase. The *rest gesture* accuracy was similar to the model trained by the first strategy as indicated in **Figure 7B**.

Compared to the classifier trained by reaching phase (the first strategy, shown in **Figure 7**) with $t_i = -729$ ms, the probability intersection of this second-strategy model appeared 395 ms later in time, at $t_i = -334$ ms. This illustrates that the EMG data of the grasping phase contained less information for detecting gestures in the earlier stage compared to the reaching-phase data, and thus the model trained by the second strategy provided less response time for the system. However, the *grasp gesture* probability peak of this second model outperformed the *top competitor* by $d_p = 0.41$, which was 0.05 higher than $d_p = 0.36$ of the first model. This demonstrates that the dynamic EMG from the grasping phase is more accurate and confident for decision-making during the grasp movement, outperforming other interference options to a larger degree.

## 3.3. Training With Reaching, Grasping, and Resting Phases

The performance of the classifier trained by the third strategy (trained with reaching, grasping, and resting phases) is shown in **Figure 9**. In **Figure 9A**, the predicted probability of *grasp gesture* increased steadily during the reach-to-grasp movement when the grasp was carried out from the resting status, then stayed at its peak during the reaching and grasping phases, and gradually decreased when the subject finished grasping and returned to resting status again. Simultaneously, the predicted probability of *rest gesture* first reduced dramatically below 0.2 during the grasp movement, until the hand returned to the rest position when the *rest gesture* probability progressively increased again. The predicted probability of the *top competitor* remained stably below 0.2, which was constantly outperformed by *grasp gesture* significantly throughout the entire trial. The intersection of *grasp gesture* and *rest gesture* probabilities happened at $t_i = -743$ ms, and the probability peak of *grasp gesture* was $d_p = 0.49$ higher than the simultaneous *top competitor*.

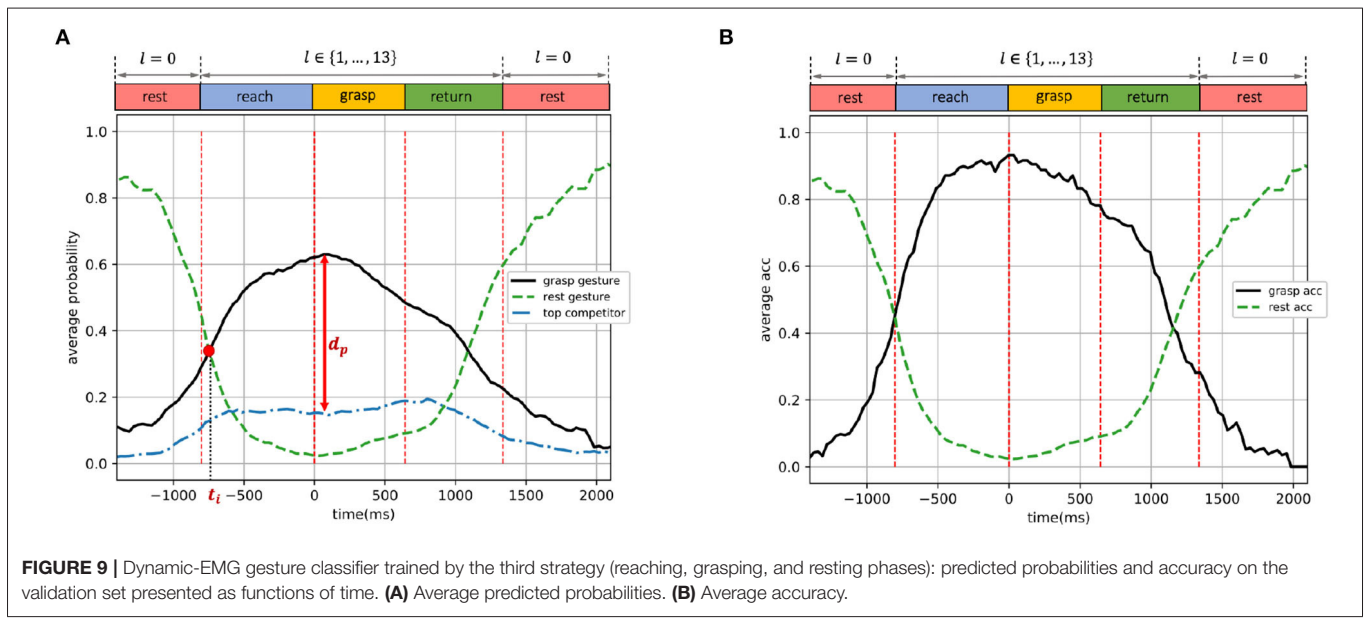

FIGURE 9 | Dynamic-EMG gesture classifier trained by the third strategy (reaching, grasping, and resting phases): predicted probabilities and accuracy on the validation set presented as functions of time. (A) Average predicted probabilities. (B) Average accuracy.

| Training Phases | $t_i$ | $d_p$ | Prob. Peak |
|---|---|---|---|
| Reach + Rest | $-729$ms | 0.36 | Reach |
| Grasp + Rest | $-334$ms | 0.41 | Grasp |
| Reach + Grasp + Rest | $-743$ms | 0.49 | Reach + Grasp |

FIGURE 10 | The summary of model performances trained by the first, second, and third training strategy.

For the dynamic gesture classification, as shown in **Figure 9B**, the average accuracy of detecting *grasp gesture* was stably higher than 0.8 within most of the reaching and grasping phases, which are the most critical phases for making a robotic-grasp decision. It is worth noting that, the validation accuracy was still higher than 0.75 at the beginning of the returning phase even though the model was not trained on any data from that phase. The average-accuracy curve of the *rest gesture* presented a similar trend as shown in **Figures 7B**, **8B**.

Compared to the first and second training strategies, the third strategy combined the advantages from both the reaching and grasping phases, leading to a boosted performance in detecting dynamic EMG compared to every single phase of reaching and grasping. The intersection point $t_i = -743$ ms moved toward the beginning of the reaching phase even further compared to the first training strategy, and simultaneously the model was able to distinguish between *grasp gesture* and *top competitor* with a higher degree of confidence ($d_p = 0.49$) throughout the entire trial compared to the second training strategy. Therefore, the model could provide even more response time before the grasp happens with more precise performance. Even during the returning phase which was unseen in the training, the model could still perform decently, illustrating the gain of using dynamic EMG for training. This higher degree of freedom in EMG data could enable more information and stability of the model to a wider range of postures during the dynamic grasp activity. In addition, the accuracy for detecting the resting phase was also highly accurate and sensitive to perform as a detector of muscle activation for triggering the robotic grasp.

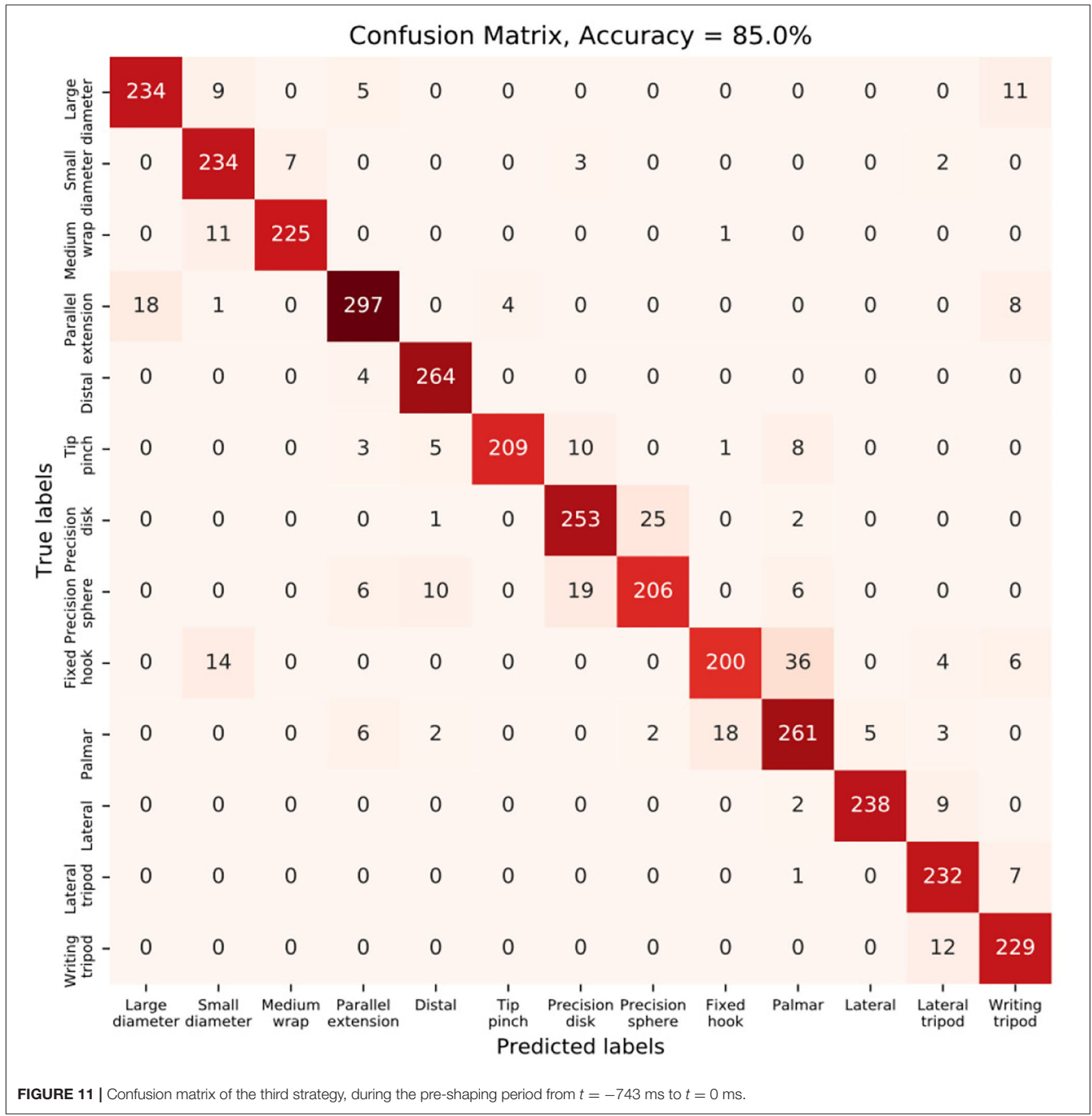

FIGURE 11 | Confusion matrix of the third strategy, during the pre-shaping period from $t = -743$ ms to $t = 0$ ms.

## 4. DISCUSSION

The summary of model performances under different classification scenarios is shown in **Figure 10**. In short, the dynamic data from the reaching phase could provide more response time for the robot to pre-shape before the grasp happens, the dynamic data from the grasping phase could improve the decision confidence and robustness compared to other competing gestures, and the higher data variability from combining the reaching and grasping phases could leverage the advantages from both phases and further boost the performance of dynamic-EMG classification.

In order to investigate the model performance regarding different gestures, in **Figure 11** we plot the confusion matrix of the model trained by the third strategy, which was evaluated on the left-out validation set. Here, we only present the evaluation result during the pre-shaping time period from $t = -743$ ms to $t = 0$ ms, in order to inspect the model's capability of detecting upcoming gestures during the reaching phase before the actual grasp happens.

As shown in **Figure 11**, the model was able to detect the upcoming gestures efficiently during the pre-shaping period, with an average accuracy of 85%. In addition, the model could also clearly distinguish between different gestures, even though it was trained with large gesture vocabularies with multiple dynamic motion positions.

But there are gestures that are very similar to each other and, thus, less distinguishable as illustrated in **Figure 11**, where groups of gestures are more likely to be misidentified from each other. For example, in **Figure 11**, the precision disk and precision sphere could be more confusing to the classifier and they were more likely to be misidentified from each other. That could be the result of similar hand configurations of those two gestures, only with few differences in the hand opening width during the reach-to-grasp movement, and the precision sphere could be identified as an intermediate state of the precision disk when the hand opens. Similar circumstances have also been observed and discussed in Sburlea and Müller-Putz (2018), where pattern similarities of the hand activities of different grasp types were extracted from three different data modalities (EEG, EMG, and joint angles) respectively. It is indicated in Sburlea and Müller-Putz (2018) that similarities of different gestures may appear in three aspects: 1. object shape (depicting the relation between grasps based on the shape and size of the grasped object), 2. grasp categorization (describing the relation between grasps based on categorization in three types, including power, precision, and intermediate), and 3. thumb position (showing the relation between grasps based on the position of the thumb including abducted or adducted). Our research conforms with this finding that gesture pairs could be less distinguishable if they are similar in any of the three aspects, as shown in **Figure 11**. We found that the grasps that involved objects with a thinner or elongated shape were confusing for the classifier to distinguish, e.g., lateral, lateral tripod, and writing tripod. It was also observed that gestures of the same grasp categorization could be less distinguishable from each other, e.g., precision disk and precision sphere. In addition, we noticed that gestures based on the same thumb position were more likely to be misidentified from each other, e.g., fixed hook and palmar.

In our future experiment, in order to better distinguish the subtle differences between similar gestures, the time period of the pre-shaping stage could be divided into finer-grained sub-phases for capturing precise changes in finger configurations.

## 5. CONCLUSION

In this article, we proposed a non-static EMG recognition method for identifying real-time hand/arm movements regarding the dynamic muscular activity variation in practice. The presented framework was trained and validated by EMG signals collected from continuous grasping tasks with variations on dynamic hand postures so that the transitions from one intent to another could be encoded into the model based on natural sequences of human grasp movements. The obtained EMG data was segmented unsupervised into different dynamic motion sequences and further labeled based on the specific motion. A classifier was then constructed for recognizing the gesture label based on the dynamic EMG signal, and the impact of different movement phases was examined *via* comparative experiments. The proposed method was assessed in a real-time manner and the corresponding performance variation over time was presented. Results illustrated the effectiveness of the framework built with the EMG data of a high degree of freedom.

## DATA AVAILABILITY STATEMENT

The raw data supporting the conclusions of this article will be made available by the authors, without undue reservation.

## ETHICS STATEMENT

Ethical review and approval was not required for the study on human participants in accordance with the local legislation and institutional requirements. The patients/participants provided their written informed consent to participate in this study.

## AUTHOR CONTRIBUTIONS

MH and MZ conducted the research and wrote the manuscript. MH, MZ, and SG organized the data collection. MH realized the methods and completed the experiments. GS and DE supervised this study and conceived the main idea of this study. All authors contributed to the article and approved the submitted version.

## FUNDING

This study is partially supported by NSF (CPS-1544895, CPS-1544636, and CPS-1544815).